\documentclass[final,1p,times]{elsarticle}

\usepackage{graphicx}
\usepackage{amssymb}
\usepackage{lineno}

\usepackage{todonotes}
\usepackage{hyperref}
\journal{Elsevier Gait \& Posture:~\url{https://doi.org/10.1016/j.gaitpost.2020.07.114}}

\begin{document}

\begin{frontmatter}


%\title{Explaining Automated Gender Classification from Human Gait}
\title{Explaining Automated Gender Classification from Human Gait}

\author[add1]{Fabian Horst\corref{contrib}}
\author[add2]{Djordje Slijepcevic\corref{contrib}}
\author[add3]{Brian Horsak}
\author[add4]{Sebastian Lapuschkin}
\author[add3]{Anna-Maria Raberger}
\author[add4]{Wojciech Samek}
\author[add5]{Christian Breiteneder}
\author[add1]{Wolfgang Immanuel Sch\"ollhorn}
\author[add2]{Matthias Zeppelzauer}

%\address[add_coauthor]{Both authors contributed equally to this research.}
\address[add1]{Department of Training and Movement Science, Institute of Sport Science, Johannes Gutenberg-University Mainz, Mainz, Germany}
\address[add2]{Institute of Creative Media Technologies, Department of Media \& Digital Technologies, St. P\"olten University of Applied Sciences, St. P\"olten, Austria}
\address[add3]{Institute of Health Sciences, Department of Health Sciences, St. P\"olten University of Applied Sciences, St. P\"olten, Austria}
\address[add4]{Department of Video Coding \& Analytics, Fraunhofer Heinrich Hertz Institute, Berlin, Germany}
\address[add5]{Institute of Visual Computing and Human-Centered Technology, TU Wien, Vienna, Austria}

%Both authors contributed equally to this research.
%\address{California, United States}

%\begin{abstract}
%\end{abstract}
%The abstract title is limited by 20 words and must be submitted using sentence case (e.g. This is the title of my abstract).

%Up to 10 authors can be submitted for an abstract (including the presenting author). The presenting author is selected first; other authors can be added only when the presenting author is submitted. However, the author order can be changed if needed by swapping the names in the list of the authors. The first name is always considered to be the main author.

%The maximum abstract length is 450 words without references.
%\end{abstract}

%\begin{keyword}
%Science \sep Publication \sep Complicated


%\end{keyword}

\cortext[contrib]{Both authors contributed equally to this research.}

\end{frontmatter}

%\linenumbers


%\todo[inline]{Der Titel sollte nicht der das gleiche Muster haben wie das ex-Frontiers paper. Das sieht später in der Publikationsliste nach Copy-Paste aus. Macht keinen guten Eindruck. Titelvorschlag z.B.: "Explaining Automated Gender Classification from Human Gait"}\todo[inline]{Vorschlag von Matthias finde ich gut, ggf. noch ändern auf ... Gender Classification in Clinical Gait Analysis / from Clinical Gait Analyis Data}
\section{Introduction}
\label{S:1}

State-of-the-art machine learning (ML) models are
highly effective in classifying gait analysis data, however, they lack in providing explanations for their predictions~\cite{halilaj2018machine}.
This "black-box" characteristic makes it impossible to
understand on which input patterns, ML models base their predictions. The present study investigates whether Explainable Artificial Intelligence methods, i.e., Layer-wise Relevance
Propagation (LRP)~\cite{bach2015pixel}, can be useful to enhance the explainability of ML predictions in gait classification.

\section{Research Question}

Which input patterns are most relevant for an automated gender classification model and do they correspond to characteristics identified in the literature?

\section{Methods}

We utilized a subset of the GAITREC dataset~\cite{horsak2020gai} containing five bilateral ground reaction force (GRF) recordings per person during barefoot walking of 62 healthy participants: 34 females (34.0$\pm$9.5 years; 65.2$\pm$11.8 kg; 4.2$\pm$0.3 m/s) and 28 males (38.5$\pm$11.9 years; 80.9$\pm$14.0 kg; 4.0$\pm$0.3 m/s). 
%- walking unassisted at self-selected walking speed on an approximately 10 m walkway with two centrally-embedded force plates (Kistler, Type 9281B12, Winterthur, CH). Data were recorded at 2000 Hz, filtered with a zero-lag Butterworth filter of 2nd order with a cut-off frequency of 20 Hz, time-normalized to 101 points (100\% stance), and amplitude-normalized to 100\% body weight. During one session subjects walked barefoot or in socks until a minimum number of 5 valid recordings were available.
Each input signal (right and left side) was min-max normalized before concatenation and fed into a multi-layer Convolutional Neural Network (CNN). 
The classification accuracy was obtained over a stratified ten-fold cross-validation.
To identify gender-specific patterns, the input relevance scores were derived using LRP.
%As an alternative perspective on the data, 1D Statistical Parametric Mapping (SPM), i.e., a two-sample t-test (p $<$ 0.05), was conducted on the whole dataset.

\section{Results}
\label{S:2}

The mean classification accuracy of the CNN with 83.3$\pm$11.4\% showed a clear superiority over the zero-rule baseline (54.8\%). Figure 1 shows averaged GRF signals with the color-coded averaged relevance scores.
%\todo[inline]{Ich glaube die Textbox in Fig1 mit den Referenzen kann man nicht lesen. Die Schriftgröße ist generell eigentlioch zu klein. Mir ist bewusst, dass das ggf. nciht wirklich besser geht, wollte euch aber dennoch darauf hinwesien ... Zumindest die x Achse könnte man größer machen, dort würde ich ggg.f die Ablürzungen sogar ausschrieben, Platz wäre dafür da.}
\begin{figure}[htb!]
  \centering	\includegraphics[width=1\linewidth]{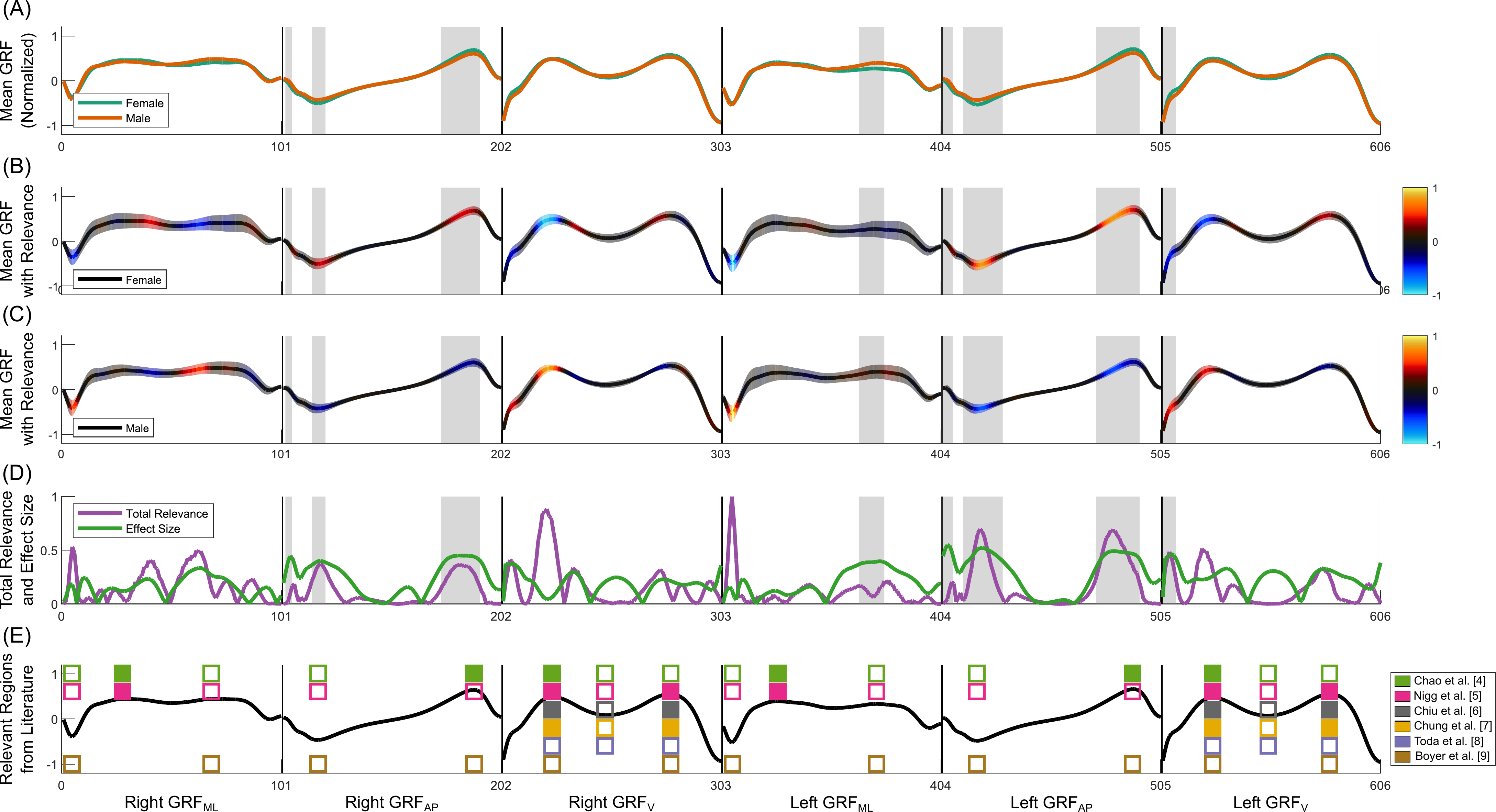} %width=0.90
    \caption{Explainability results for the gender classification. (A) Averaged GRF signals for both classes. 
    %The first three signals represent the three GRF components of the right side and are followed by the three GRF components of the left side. 
    %Note that the data for both sides is composed of three GRF components (e.g., input features of the right side: 1 to 101 ($F_{ML}$), 102 to 202 ($F_{AP}$), and 203 to 303 ($F_{V}$)). 
    %This means, for example, that input features 21 ($F_{ML}$), 122 ($F_{AP}$) and 233 ($F_{V}$) all correspond to the relative time of 20\% of the same stance phase. 
    The shaded areas differ significantly (two-sample t-test, p$<$0.05) according to 1D Statistical Parametric Mapping (SPM).
    %The shaded areas highlight the input signals where 1D Statistical Parametric Mapping (SPM), i.e., a two-sample t-test (p $<$ 0.05), resulted in a statistically significant difference between both classes. 
    (B)-(C) Averaged GRF signals for female/male class, with a band of one standard deviation, color-coded via LRP relevance scores. 
    %for the $F$ class obtained using LRP. (C) Averaged GRF signals of all test trials as a line plot for the $M$ class, in the same format as in (B). 
    (D) Effect size obtained from SPM and total relevance (absolute sum of input relevance scores of both classes).
    %The total relevance indicates the common relevance of the input signal for the classification task. 
    (E) Significant (filled boxes) and non-significant (empty boxes) discrete GRF characteristics according to the literature~\cite{chao1983normative,nigg1994gait,chiu2007effect, chung2010change, toda2015age, boyer2008gender}.}
    \label{img:cnn-norm}
\end{figure}

\section{Discussion}
%The relatively high classification accuracy, as well as the high deviation from the Zero-rule Baseline, allows a robust analysis of the explainability results. 
According to LRP, relevant regions reside in all GRF signals. Most relevant regions include the first peak and a section during late stance of the $GRF_{ML}$, the first peak of $GRF_{V}$, and both peaks in $GRF_{AP}$. To interpret the LRP results, we compared them with (I) SPM and (II) the literature. (I) The LRP regions in $GRF_{AP}$ differ significantly between the two classes, according to SPM. This does not apply to LRP regions in $GRF_{V}$ and $GRF_{ML}$. (II) The regions around the first $GRF_{V}$ peak and the second $GRF_{AP}$ peak match with discrete characteristics reported in the literature. An additional correspondence with the second $GRF_{V}$ peak was observed in a separate experiment where only $GRF_{V}$ was used for classification.
%In contrast to discrete characteristics that have been shown to differ between women and men in the literature, the first peak of $F_{V}$ is the only common observation with the explainability results. 
%The first peak of $F_{V}$ represents the only overlap with the literature.
%A comparison with the literature shows that only the first peak value of $F_{V}$ overlaps.
Overall, the results of LRP, SPM, and the literature are partly consistent.
%LRP and the two reference methods highlight different aspects in the data, or that the composition of the sample or other influencing factors (such as age) explain those differences.
%However, the explainability results show that LRP identifies regions of the GRF signals that are relevant for predictions of an ML model and represents a valuable tool for a better understanding of ML predictions.
%\todo[inline]{Der Abschlussatz war mir zu generisch und zuwenig verbunden mit dem Paper - hier mein neuer Vorschlag:}
Our results suggest that ML models ground their predictions on different information than identified in the literature and through statistical analysis. This raises interesting questions for further research.
%\todo[inline]{Mein Vorschlag zu Matthias: Based on our results it might be concluded, that ML models ground their decisions on different information than identified in the literature and through statistical analysis. This raises interesting questions for further research on this topic. Fbaia Djordje: durch die Comments konnte ich jetzt leider nciht die Wortanzahl im Auge behalten, sry. Aber ggf. lässt sich mal bei den demographsichen Daten oben in der Methdok etwas sparen.}

%Nevertheless, this experiment is a first step, and further experiments with larger data are necessary to better understand gender-specific gait patterns.

%\section{References}

\section*{Funding}
This work was partly funded by the Austrian Research Promotion Agency (FFG) and the BMDW within the COIN-program (\#866855 and \#866880),
the Lower Austrian Research and Education Company (NFB), the Provincial Government of Lower Austria (\#FTI17-014).
Further support was received from the German Ministry for Education and Research as BIFOLD (\#01IS18025A and \#01IS18037A) and TraMeExCo (\#01IS18056A).

\section*{Acknowledgments}
We want to thank Marianne Worisch, Szava Zolt\'{a}n, and Theresa Fischer for their great assistance in data preparation and their support in clinical and technical questions.

%% New version of the num-names style
\bibliographystyle{elsarticle-num-names}
\bibliography{sample.bib}






% \end{thebibliography}

\end{document}